\documentclass
[
     preprint
]{article}

\usepackage[nonatbib]{nips_2018}

\usepackage[utf8]{inputenc} 
\usepackage[T1]{fontenc}    
\usepackage{url}            
\usepackage{booktabs}       
\usepackage{amsfonts}       
\usepackage{nicefrac}       
\usepackage{microtype}      

\usepackage[pdffitwindow=false,
            plainpages=false,
            pdfpagelabels=true,
            pdfpagemode=UseOutlines,
            pdfpagelayout=SinglePage,
            bookmarks=false,
            colorlinks=true,
            hyperfootnotes=false,
            linkcolor=blue,
            urlcolor=blue!30!black,
            citecolor=green!50!black]{hyperref}

\usepackage[framemethod=TikZ]{mdframed} 

\usepackage
{
    graphicx,
    amssymb,
    amsmath,
    amsthm,
    dsfont, 
    xcolor,
    enumitem,
    mathtools,
    ifthen,
}

\newtheorem{theorem}{Theorem}[section]

\newtheorem{definition}[theorem]{Definition}
\theoremstyle{definition}
\newtheorem{example}[theorem]{Example}

\makeatletter
\def\thm@space@setup{\thm@preskip=2pt
\thm@postskip=0pt}
\makeatother

\newcommand{\ts}{\hspace*{0.1em}} 

\newcommand{\subfiguretitle}[1]{{\scriptsize{#1}} \\[0.25ex]}
\newcommand{\R}{\mathbb{R}}                                     
\newcommand{\innerprod}[2]{\left\langle #1,\, #2 \right\rangle} 
\newcommand{\dd}{\mathrm{d}}                                    
\providecommand{\norm}[1]{\left\lVert #1 \right\rVert}          

\newcommand\restr[2]{{\left.\kern-\nulldelimiterspace #1 \vphantom{\big|} \right|_{#2}}} 

\newcommand\xqed[1]{\leavevmode\unskip\penalty9999 \hbox{}\nobreak\hfill \quad\hbox{#1}}
\newcommand{\exampleSymbol}{\xqed{$\blacktriangle$}} 

\newcommand{\inspace}{\mathbb{X}} 

\newcommand{\idop}{\mathcal{I}} 

\newcommand{\ebd}[1][]{
   \ifthenelse{\equal{#1}{}}{\mathcal{E}}{\mathcal{E}_{#1}}}
\newcommand{\pro}[1][]{
   \ifthenelse{\equal{#1}{}}{\mathcal{Q}}{\mathcal{Q}_{#1}}}

\newcommand{\pf}[1][]{
   \ifthenelse{\equal{#1}{}}{\mathcal{P}}{\mathcal{P}_{#1}}}
\newcommand{\epf}[1][]{
   \ifthenelse{\equal{#1}{}}{\widehat{\mathcal{P}}}{\widehat{\mathcal{P}}_{#1}}}
\newcommand{\ko}[1][]{
   \ifthenelse{\equal{#1}{}}{\mathcal{K}}{\mathcal{K}_{#1}}}
\newcommand{\eko}[1][]{
   \ifthenelse{\equal{#1}{}}{\widehat{\mathcal{K}}}{\widehat{\mathcal{K}}_{#1}}}

\newcommand{\cov}[1][]{\mathcal{C}_\mathit{\scriptscriptstyle #1}} 
\newcommand{\ecov}[1][]{\widehat{\mathcal{C}}_\mathit{\scriptscriptstyle #1}} 

\newcommand{\gram}[1][]{G_\mathit{\scriptscriptstyle #1}} 

\mathtoolsset{centercolon}

\allowdisplaybreaks

\title{Analyzing high-dimensional time-series data using kernel transfer operator eigenfunctions}

\author{
    Stefan Klus \\
    Department of Mathematics and Computer Science \\
    Freie Universit\"at Berlin \\
    14195 Berlin, Germany \\
    \texttt{stefan.klus@fu-berlin.de}
    \And
    Sebastian Peitz \\
    Department of Mathematics \\
    Paderborn University \\
    33098 Paderborn, Germany \\
    \texttt{speitz@math.upb.de}
    \And
    Ingmar Schuster \\
    Department of Mathematics and Computer Science \\
    Freie Universit\"at Berlin \\
    14195 Berlin, Germany \\
    \texttt{ingmar.schuster@fu-berlin.de}
}

\begin{document}

\maketitle

\begin{abstract}
Kernel transfer operators, which can be regarded as approximations of transfer operators such as the Perron--Frobenius or Koopman operator in reproducing kernel Hilbert spaces, are defined in terms of covariance and cross-covariance operators and have been shown to be closely related to the conditional mean embedding framework developed by the machine learning community. The goal of this paper is to show how the dominant eigenfunctions of these operators in combination with gradient-based optimization techniques can be used to detect long-lived coherent patterns in high-dimensional time-series data. The results will be illustrated using video data and a fluid flow example.
\end{abstract}

\section{Introduction}

Hidden Markov models (HMMs) and particle filtering methods are a mainstay of machine learning and statistics for sequential and time series data \cite{RoGha1999}. They have originated in dynamical systems research and then developed independently. In this paper, we adapt and extend recent advances in dynamical systems theory, in particular kernel transfer operators, to the machine learning and statistics setting in order to analyze sequential data sets. The approximation of transfer operators and their eigenfunctions has important applications in molecular dynamics \cite{NKPMN14, SP15, KKS16}, fluid dynamics \cite{RMBSH09, Schmid10, TRLBK14}, control theory \cite{KKB17, KoMe18, Pei18}, and many other areas. Over the last years, data-driven techniques gained considerable popularity. Based only on simulation or measurement data, dominant dynamics or modes can be extracted, which are then, for instance, used for dimensionality reduction, the detection of metastable sets, system identification, or control. A similar concept has been proposed by the machine learning community. Probability densities are embedded into reproducing kernel Hilbert spaces and mapped forward by the conditional mean embedding operator \cite{Song2013, MFSS16}, which can be regarded as a Perron--Frobenius operator for embedded densities. It was shown in \cite{KSM17} that eigenfunctions of transfer operators and embedded transfer operators can be estimated simply by solving an eigenvalue problem involving (time-lagged) Gram matrices. The resulting methods are closely related to kernel Extended Dynamic Mode Decomposition (EDMD) and its variants \cite{WRK15, SP15}. As a result, it is possible to apply these methods to non-vectorial data such as strings or graphs or any other domain where a similarity measure given by a kernel is available. We will apply methods to estimate eigenfunctions of transfer operators to high-dimensional time-series data such as videos and fluid dynamics simulations and demonstrate how to extract information about the slow dominant processes of the underlying system. Our main contributions are as follows: We first present an approach to summarize the dynamical nature of sequential data in single hypothetical snapshots by solving an optimization problem. Furthermore, we highlight the connections between HMMs, change-point detection, and transfer operators and demonstrate their ability to find critical time points in sequential data beyond change points.

\section{Prerequisites}

In this section, we will briefly introduce transfer operators \cite{LaMa94, DJ99, BMM12, KNKWKSN18}, reproducing kernel Hilbert spaces \cite{Schoe01, StCh08}, and kernel transfer operators \cite{KSM17}. For more details, see the references above.

\subsection{Transfer operators}

Let $ T \colon \inspace \to \inspace $ be a discrete dynamical system, where $ \inspace \subseteq \R^d $. Furthermore, let $ p(y \mid x) $ be the \emph{transition density function}. That is, $ p(y \mid x) $ is the probability that $ X_{k+1} = y $ given that $ X_k = x $.

\begin{definition}
The \emph{Koopman operator} $ \ko $ and the \emph{Perron--Frobenius operator} $ \pf $ with respect to the measure $ \mu $ are defined by
\begin{equation*}
    \setlength{\belowdisplayskip}{0pt}
    \ko f(x) = \int p(y \mid x) \ts f(y) \ts \dd \mu(y)
    \quad \text{and} \quad 
    \pf f(y) = \int p(y \mid x) \ts f(x) \ts \dd \mu(x).
\end{equation*}
\end{definition}

It follows that $ \innerprod{\pf f}{g}_\mu = \innerprod{f}{\ko g}_\mu $. The Koopman operator can also be written as $ \ko f(x) = \mathbb{E}[f(T(x))] $ or, if the system is deterministic, simply as $ \ko f(x) = f(T(x)) $. The eigenvalues and eigenfunctions of these operators contain global information about characteristic properties of the system such as metastable sets and their time scales.

\subsection{Covariance operators}

In what follows, $ k $ denotes a positive definite kernel on $ \inspace \times \inspace $ and $ \mathbb{H} $ the corresponding reproducing kernel Hilbert space with inner product $ \innerprod{\cdot}{\cdot}_\mathbb{H} $. That is, it holds that $ \innerprod{f}{k(x, \cdot)}_\mathbb{H} = f(x) $ for all $ f \in \mathbb{H} $. This implies that $ \innerprod{k(x, \cdot)}{k(x^\prime, \cdot)}_\mathbb{H} = k(x, x^\prime) $. Moreover, $ k(x, \cdot) $ can be interpreted as the feature map $ \phi(x) $ of $ x $ so that $ k(x, x^\prime) = \innerprod{\phi(x)}{\phi(x^\prime)}_\mathbb{H} $. With these definitions, we can now introduce covariance operators. Since the dynamical system $ T $ is a mapping from $ \inspace $ to itself, it suffices for our purposes to consider the case where the input space and output space are identical. The slightly more general setting where the two spaces are different can be found in~\cite{Baker1973, MFSS16}.

\begin{definition}
The \emph{covariance operator} $ \cov[XX] \colon \mathbb{H} \to \mathbb{H} $ and \emph{cross-covariance operator} $ \cov[YX] \colon \mathbb{H} \to \mathbb{H} $ are defined as
\begin{equation*}
    \setlength{\belowdisplayskip}{0pt}
    \cov[XX] = \int \phi(X) \otimes \phi(X) \ts \dd \mu(X)
    \quad \text{and} \quad
    \cov[YX] = \int \ko \phi(X) \otimes \phi(X) \ts \dd \mu(X).
\end{equation*}
\end{definition}

Since the covariance and cross-covariance operator can in general not be computed analytically, we estimate these operators from training data. Given $ \mathbb{D}_\mathit{\scriptscriptstyle XY} = \{ (x_1, y_1), \dots, (x_n, y_n)\} $, where $ y_i = T(x_i) $, we define the data matrices $ \mathbf{X} = [x_1, \dots, x_n] $ and $ \mathbf{Y} = [y_1, \dots, y_n] $ and the feature matrices $ \Phi = [\phi(x_1), \dots, \phi(x_n)] $ and $ \Psi = [\phi(y_1), \dots,\phi(y_n)] $ so that the empirical estimates of $ \cov[XX] $ and $ \cov[YX] $ can be computed as
\begin{equation*}
    \ecov[XX] = \frac{1}{n} \Phi \Phi^\top
    \quad \text{and} \quad
    \ecov[YX] = \frac{1}{n} \Psi \Phi^\top.
\end{equation*}
Furthermore, we define the Gram matrix $ \gram[XX] $ and time-lagged Gram matrices $ \gram[XY] $ and $ \gram[YX] $ by
\begin{equation*}
    \gram[XX] = \Phi^\top \Phi, \quad \gram[XY] = \Phi^\top \Psi, \quad \text{and} \quad \gram[YX] = \Psi^\top \Phi.
\end{equation*}
The goal is to reformulate the eigenvalue problems derived below involving covariance and cross-covariance matrices in terms of Gram matrices and time-lagged Gram matrices.

\subsection{Kernel transfer operators}

We assume that $ \cov[XX]^{-1} $ exists or otherwise replace it by its regularized version $ (\cov[XX] + \tilde{\varepsilon} \ts \idop)^{-1} $, where $ \tilde{\varepsilon} $ is a regularization parameter and $ \idop $ the identity operator.

\begin{definition}
The kernel Koopman operator~$ \ko[k] $ and the kernel Perron--Frobenius operator $ \pf[k] $ are defined as
\begin{equation*}
    \ko[k] = \cov[XX]^{-1} \ts \cov[XY]
    \quad \text{and} \quad
    \pf[k] = \cov[XX]^{-1} \ts \cov[YX].
\end{equation*} 
\end{definition}

Thus, in order to compute eigenfunctions of these operators, we have to solve
\begin{equation*}
    \cov[XX]^{-1} \ts \cov[XY] \ts \varphi = \lambda \ts \varphi
    \quad \text{and} \quad
    \cov[XX]^{-1} \ts \cov[YX] \ts \varphi = \lambda \ts \varphi,
\end{equation*}
respectively, or the corresponding eigenvalue problems where the covariance and cross-covariance operators are replaced by their empirical estimates $ \ecov[XX] $, $ \ecov[XY] $, and $ \ecov[YX] $. The Perron--Frobenius operator is then approximated with respect to the measure given by the density of the training data $ \textbf{X} $. Information about the slow processes of the system, encoding its most important dynamical properties, is given by the eigenfunctions associated with eigenvalues close to one, see \cite{KNKWKSN18}. Eigenvalues are thus sorted in decreasing order. It was shown in \cite{KSM17} that eigenfunctions $ \varphi $ can be approximated using only kernel function evaluations (see also \cite{WRK15, SP15}). This is summarized in the following algorithms:

\begin{minipage}{0.49\textwidth}
    \begin{mdframed}[roundcorner=4pt]
        \centering \textbf{Koopman operator}
        \begin{enumerate}[leftmargin=1em]
        \item Compute Gram matrices $ \gram[XX] $ and $ \gram[YX] $.
        \item Solve $ \big(\gram[XX] + \varepsilon I\big)^{-1} \ts \gram[YX] \ts v = \lambda \ts v $.
        \item Set $ \varphi = \Phi \ts v $.
        \end{enumerate}
    \end{mdframed}
\end{minipage}
\hfil
\begin{minipage}{0.49\textwidth}
    \begin{mdframed}[roundcorner=4pt]
        \centering \textbf{Perron--Frobenius operator}
        \begin{enumerate}[leftmargin=1em]
        \item Compute Gram matrices $ \gram[XX] $ and $ \gram[XY] $.
        \item Solve $ \big(\gram[XX] + \varepsilon I\big)^{-1} \gram[XY] v = \lambda v $.
        \item Set $ \varphi = \Phi \ts \gram[XX]^{-1} \ts v $.
        \end{enumerate}
    \end{mdframed}
\end{minipage}

Here, $ \varepsilon $ is a regularization parameter and $ I $ denotes the identity matrix. Let us illustrate these algorithms with the aid of a simple guiding example.

\begin{example}
Consider the one-dimensional stochastic differential equation
\begin{equation*}
    \dd X_t = -\nabla V(X_{t}) \ts \dd t + \sqrt{2 D} \, \dd W_t,
\end{equation*}
with the potential $ V $ shown in Figure~\ref{fig:TripleWell}(a), $ D = 0.28125 $, and a one-dimensional standard Wiener process $ W_t $. The system can be interpreted as a particle in an energy landscape. The particle will typically stay for a long time in one of the wells, jump with a low probability to one of the other wells, and so on. We choose the lag time $ \tau = 10 $ and analyze the resulting discrete dynamical system. Figure~\ref{fig:TripleWell}(b) shows a typical trajectory. Of particular interest are now the metastable sets corresponding to the wells. The dominant eigenfunctions of the Koopman and Perron--Frobenius operator are shown in Figure~\ref{fig:TripleWell}(c) and \ref{fig:TripleWell}(d), respectively. The first eigenfunction of the Koopman operator corresponding to $ \lambda = 1 $ is the constant function and does not contain any useful information. The second eigenfunction is almost constant within the left well and also within the two other wells with a smooth transition in between, pinpointing the highest energy barrier. The third eigenfunction separates the middle well and the right well. The eigenfunctions of the Perron--Frobenius operator contain the same information, just weighted by the invariant density. \exampleSymbol

\begin{figure}[tb]
    \centering
    \begin{minipage}[t]{0.36\textwidth}
        \centering
        \subfiguretitle{(a)}
        \includegraphics[width=\textwidth]{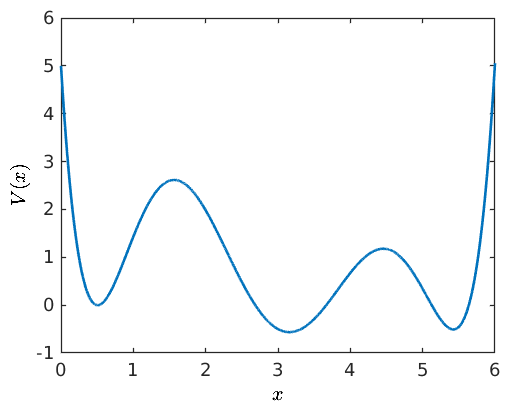}
    \end{minipage}
    \hspace*{2ex}
    \begin{minipage}[t]{0.36\textwidth}
        \centering
        \subfiguretitle{(b)}
        \includegraphics[width=\textwidth]{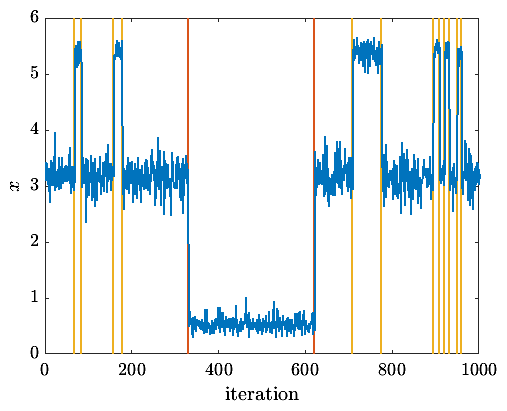}
    \end{minipage}
    \\[1ex]
    \begin{minipage}[t]{0.36\textwidth}
        \centering
        \subfiguretitle{(c)}
        \includegraphics[width=\textwidth]{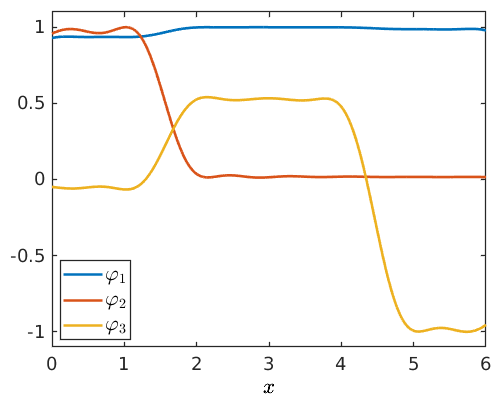}
    \end{minipage}
    \hspace*{2ex}
    \begin{minipage}[t]{0.36\textwidth}
        \centering
        \subfiguretitle{(d)}
        \includegraphics[width=\textwidth]{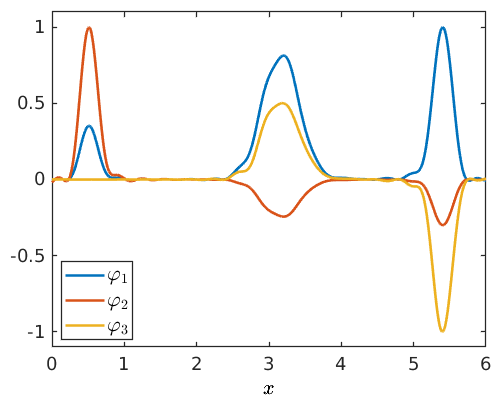}
    \end{minipage}
    \caption{(a) Triple-well potential. (b) Typical trajectory and change point detection using Koopman operator eigenfunctions (see Section~\ref{sec:Data summarization}), changes detected by the first nontrivial eigenfunction are marked in red and changes detected by the second in yellow. (c) Dominant eigenfunctions of the Koopman operator. (d) Dominant eigenfunction of the Perron--Frobenius operator with respect to the uniform density.}
    \label{fig:TripleWell}
\end{figure}

\end{example}

\section{Single snapshot data summarization}
\label{sec:Data summarization}

The example illustrates that it is possible to decompose the state space into metastable sets by considering the extrema or plateaus of the eigenfunctions. For the one-dimensional problem, it is easy to find these plateaus since we can sample the entire state space. For high-dimensional systems, we cannot explore the state space due to the extremely large number of degrees of freedom. However, given an initial guess $ x^{(0)} $, we can find better solutions by applying optimization methods. While methods like Bayesian optimization \cite{Sn12} might be applicable, we choose simple gradient descent and ascent approaches. That is, we compute
\begin{equation*}
    x^{(\ell+1)} = x^{(\ell)} \mp \eta \ts \nabla \varphi\big(x^{(\ell)}\big)
\end{equation*}
with the learning rate $ \eta $ until no local improvement can be found. If a full update step does not improve the cost function, the parameter $ \eta $ is changed adaptively. Since $ \varphi = \Phi \ts \alpha $, where $ \alpha $ is simply a vector of coefficients given by $ \alpha = v $ for the Koopman operator or $ \alpha = \gram[XX]^{-1} v $ for the Perron--Frobenius operator, we can write
\begin{equation*}
    \varphi(x) = \sum_{i=1}^n \alpha_i \ts k(x, x_i)
    \quad \text{and} \quad
    \nabla \varphi(x) = \sum_{i=1}^n \ts \alpha_i \nabla k(x, x_i),
\end{equation*}
where the gradient is computed with respect to the variable $ x $. For the Gaussian kernel, for instance, which we will use for the examples below, we obtain
\begin{equation*}
    k(x, x_i) = \exp\left(-\frac{1}{2\sigma^2} \norm{x-x_i}^2\right)
    \quad \text{and} \quad
    \nabla k(x, x_i) = -\frac{1}{\sigma^2} (x - x_i) \ts k(x, x_i).
\end{equation*}
If the globally optimal solutions are found, these correspond to the different metastable sets of the system. However, we will, in general, not find the global but only local extrema. This clearly also depends on the chosen kernel. Nevertheless, for the high-dimensional examples below, the gradient-based optimization techniques result in meaningful and interpretable solutions and represent data summaries in single (hypothetical) snapshots. By evaluating the dominant eigenfunctions for each data point of a given trajectory, it is also possible to identify change points simply by observing jumps between the minima and maxima as illustrated in Figure~\ref{fig:TripleWell}(b). If the eigenfunction value suddenly changes, this indicates a transition from one metastable set to another. Moreover, we can identify an ordering of change points based on their inherent time scales given by the eigenvalues. The possibility to detect change points is a byproduct of the spectral decomposition of transfer operators.

\section{Related work}

We now briefly discuss how transfer operator methods are related to modern time series models and associated inference methods in machine learning and statistics as well as change point detection.

\paragraph{Probabilistic time series models.}

The prototype of time series models is the hidden Markov model. The HMM assumes a hidden state that evolves in time according to some transition density $ p(y \mid x) $. The observation $ z $ at each time step depends on the hidden state through an observation density $ p_o(z \mid x) $. The Koopman operator on the other hand assumes randomness only in the hidden state, not in the observation, which is typically thought of as a deterministic function of the hidden state. Similarly, whenever no analytical solution for inference in the statistical model exists, one typically has to resort to optimization or sampling methods. Inference in the kernel transfer operator setting often starts out by an analytically given operator decomposition. Applications like change point detection and data summarization follow from this decomposition.

\paragraph{Determinantal point processes.}

Determinantal point processes (DPPs) are probabilistic models of repulsion that have been studied in random matrix theory and machine learning. Their application to machine learning problems is mainly in data summarization. Given a set $\mathcal{X}$ (for example a set of empirical observations), a positive definite kernel $k$, and a randomly drawn subset $\mathcal{S} \subseteq \mathcal{X}$, a DPP is defined as the distribution that satisfies $\mathbb{P}(\mathcal{A} \subseteq \mathcal{S}) = \det \gram[\mathcal{A}]$ for all $\mathcal{A} \in \mathcal{X}$, where $\gram[\mathcal{A}]$ is the gram matrix of $\mathcal{A}$ computed using $k$. Diverse subsets of  $\mathcal{X}$ induce larger determinants and are thus more probable under a DPP. This has previously been used for kernel-based data summarization, see \cite{Kulesza2012:Determinantal} for a review. The cost of our method for data summarization is comparable to that of a DPP, as the computation of the determinant is cubic in the number of data points. A major advantage of our approach is that it is possible to represent data summaries in hypothetical snapshots, whereas previous work on data summarization using Determinantal Point Processes~(DPPs) only chooses among previously observed snapshots. On the other hand, unlike our method, DPPs can be used for summarizing non-sequential data.

\paragraph{Change point detection.}

Change point detection is a method to identify changes in stochastic processes $\{X_1,\dots,X_T\}$ that are typically assumed to be globally and piecewise stationary. Concretely, if $t_1<t_2<\dots<t_K$ are the (unknown) change points, then we assume that $X_{t_n - 1} \sim p_{n-1}$ and $X_{t_n} \sim p_{n}$ with $p_{n-1} \neq p_{n}$. Without knowing the distributions $p_*$ or $K$, our aim is to identify $\{t_1,\dots,t_K\}$. For an in-depth overview of existing methods, see \cite{Tru2018}. The closest previous work on change point detection in the literature uses shifts in the kernel mean embedding of points in the time series to identify change points \cite{Har2007}. This is based on the elementary observation that for characteristic kernels, there is a one-to-one correspondence between distributions and kernel mean embeddings. Thus, if the kernel mean embeddings differ, i.e., $\mu_{p_{n-1}} \neq \mu_{p_n}$, this implies that $p_{n-1} \neq p_{n}$. The complexity of the procedure is $\mathcal{O}(KT^2)$, where $K$ is the number of change points and $T$ the number of measurements, whereas our approach has complexity $\mathcal{O}(T^3)$. In contrast to our method, however, there is no natural ordering of the importance of change points, and once the eigendecomposition of the kernel transfer operator is computed, our method is more versatile as it can also be used, e.g., for data summarization.

\section{Applications}
\label{sec:Applications}

In this section, we will show how to apply the proposed kernel-based methods to high-dimensional time-series data.

\subsection{Pendulum}

\begin{figure}[htb]
    \centering
    \begin{minipage}[t]{0.305\textwidth}
        \centering
        \subfiguretitle{(a)}
        \vspace*{0.5ex}
        \includegraphics[width=\textwidth]{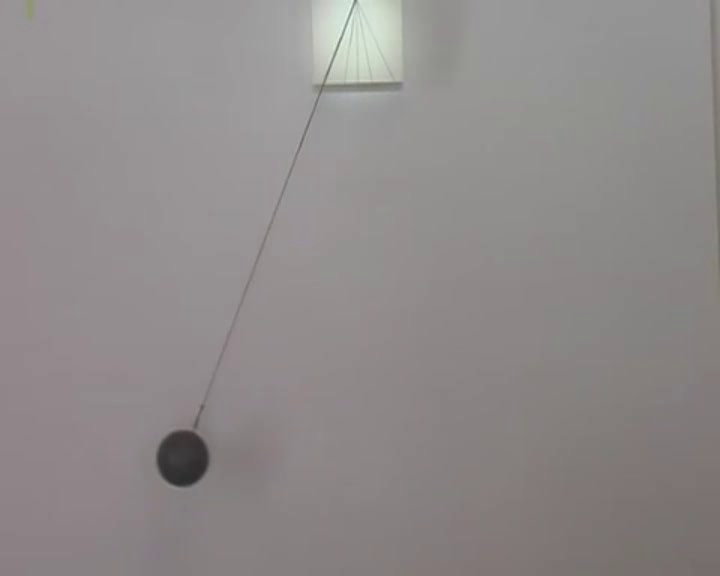}
    \end{minipage}
    \hspace*{0.5ex}
    \begin{minipage}[t]{0.285\textwidth}
        \centering
        \subfiguretitle{(b)}
        \includegraphics[width=\textwidth]{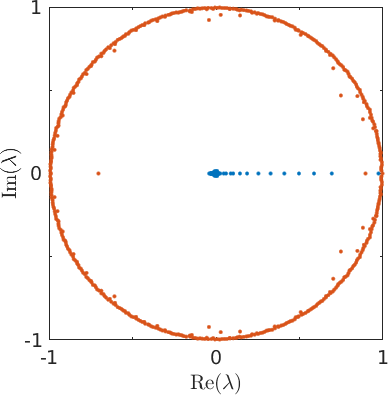}
    \end{minipage}
    \hspace*{0.5ex}
    \begin{minipage}[t]{0.35\textwidth}
        \centering
        \subfiguretitle{(c)}
        \includegraphics[width=\textwidth]{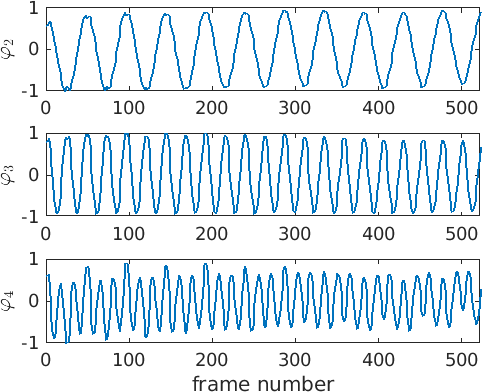} 
    \end{minipage}
    \caption{(a) First frame of the video. (b) Kernel Koopman operator eigenvalues (blue) and exact DMD eigenvalues (red). (c) Dominant kernel Koopman eigenfunctions evaluated at the frames of the video.}
    \label{fig:Pendulum}
\end{figure}

\begin{figure}[tb]
    \centering
    \begin{minipage}{0.26\textwidth}
        \centering
        \subfiguretitle{(a) $ \lambda_2 \approx 0.69 $}
        \includegraphics[width=\textwidth]{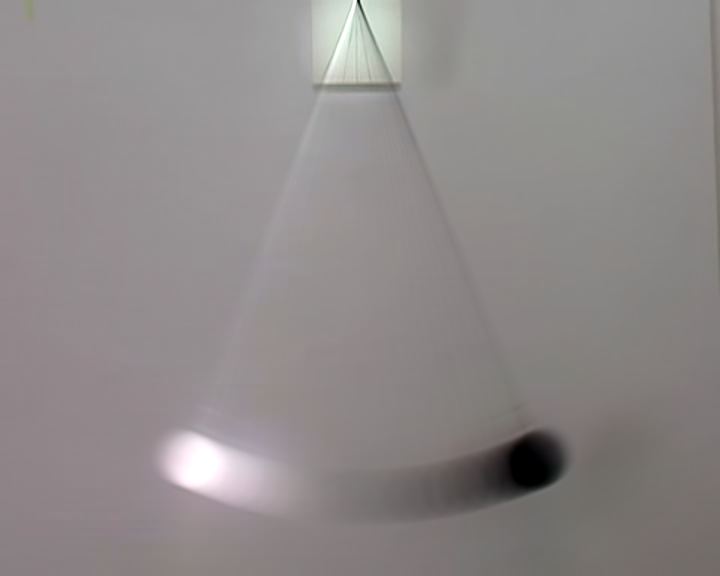} \\[0.5ex]
        \includegraphics[width=\textwidth]{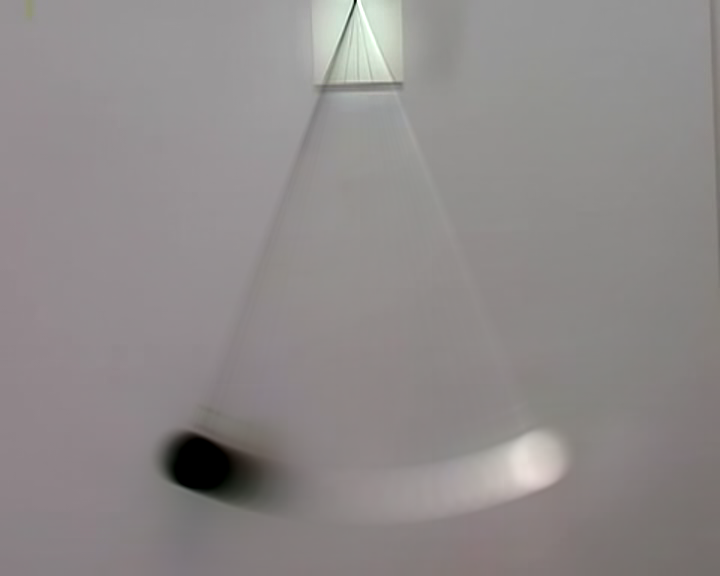}
    \end{minipage}
    \begin{minipage}{0.26\textwidth}
        \centering
        \subfiguretitle{(b) $ \lambda_3 \approx 0.58 $}
        \includegraphics[width=\textwidth]{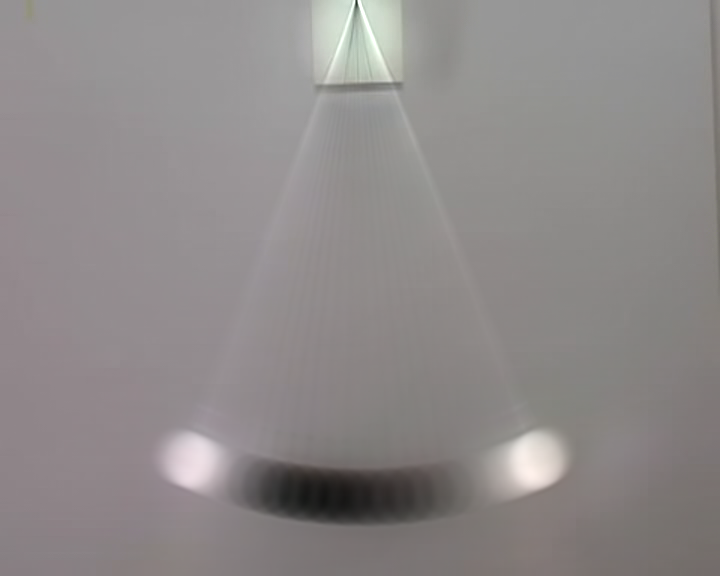} \\[0.5ex]
        \includegraphics[width=\textwidth]{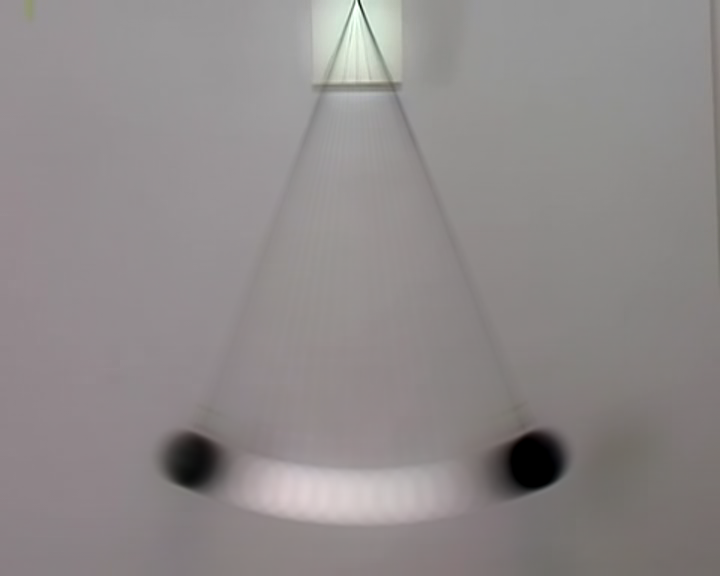}
    \end{minipage}
    \begin{minipage}{0.26\textwidth}
        \centering
        \subfiguretitle{(c) $ \lambda_4 \approx 0.49 $}
        \includegraphics[width=\textwidth]{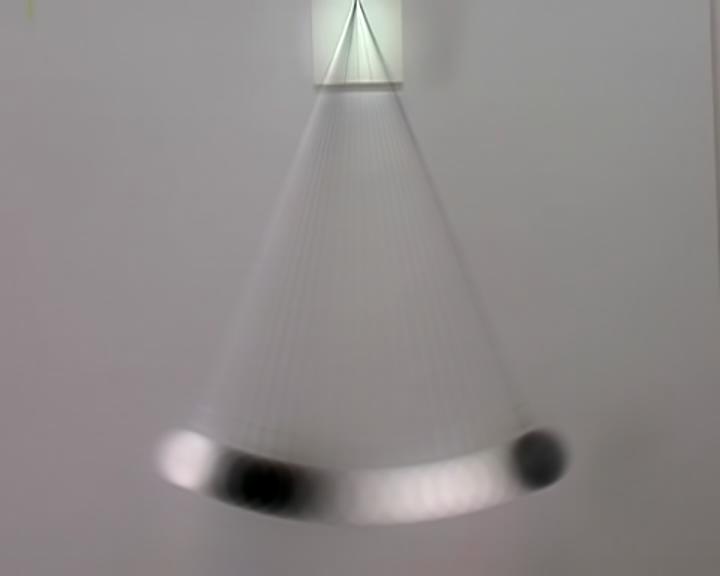} \\[0.5ex]
        \includegraphics[width=\textwidth]{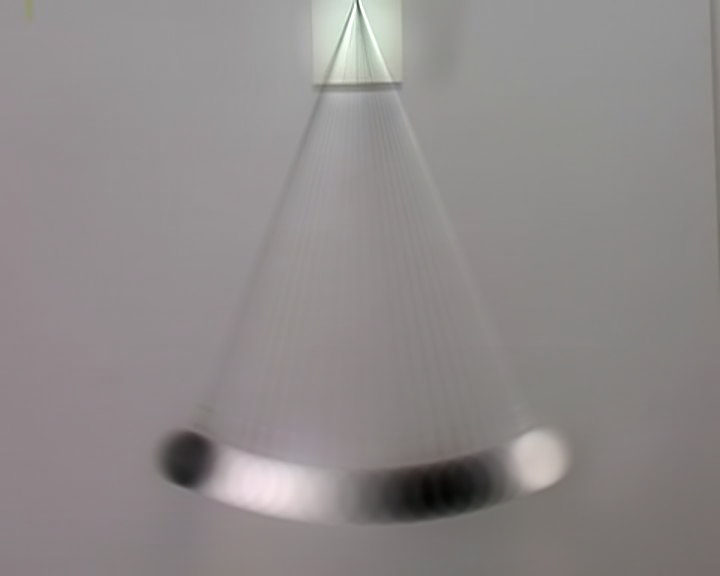}
    \end{minipage}
    \caption{Minima (top) and maxima (bottom) of the leading eigenfunctions of the kernel Koopman operator computed using gradient descent/ascent methods.}
    \label{fig:Pendulum eigenfunction maximizers}
\end{figure}

Let us illustrate the efficacy of the kernel-based methods using the example of video data, which is inherently high-dimensional. We first analyze a video showing a pendulum.\footnote{ScienceOnline: The Pendulum and Galileo (\url{www.youtube.com/watch?v=MpzaCCbX-z4}).} This example was introduced in \cite{KSM17}, where it was shown that the leading eigenfunctions evaluated at the frames of the video encode the resonant frequency and multiples of this frequency. We now want to compute the frames maximizing and minimizing these eigenfunctions using the gradient descent and ascent approaches described above. Here, we do not convert the video frames to grayscale images but directly work with the RGB values, i.e., each frame $ x $ is a tensor in $ \R^{576 \times 720 \times 3} $. The video comprises 523 frames so that $ \mathbf{X}, \mathbf{Y} \in \R^{576 \times 720 \times 3 \times 522} $ are tensors of order 4. We define the kernel by
\begin{equation*}
    k(x, x^\prime) = \exp\left(-\frac{1}{2 \sigma^2}\norm{x - x^\prime}_F^2\right)
\end{equation*}
and choose the bandwidth $ \sigma = 1.25 \cdot 10^4 $ and regularization parameter $ \varepsilon = 1 $. The first frame of the video is shown in Figure~\ref{fig:Pendulum}(a). While the eigendecomposition of the kernel Koopman operator detects several real-valued eigenvalues, Dynamic Mode Decomposition (DMD) \cite{Schmid10, TRLBK14} results mainly in complex eigenvalues as illustrated in Figure~\ref{fig:Pendulum}(b) and the associated DMD modes are harder to interpret. The dominant eigenfunctions of the kernel Koopman operator encode the frequencies of the system as shown in Figure~\ref{fig:Pendulum}(c). Note, however, that the regularization parameter $ \varepsilon $ perturbs the eigenvalues.
In addition to the spectrum and the eigenfunction values of each frame, we can now compute frames that minimize and maximize these eigenfunctions. During the gradient descent and ascent, we restrict all values to be in $ [0, 255] $ so that the resulting optimized frames can still be interpreted as images and easily visualized. The results are shown in Figure~\ref{fig:Pendulum eigenfunction maximizers}. The top and bottom images can be regarded as the maximally different states with respect to the corresponding eigenfunction. The first nontrivial dominant eigenfunction distinguishes between maximum displacement left and maximum displacement right, followed by higher-order modes. This simple example shows that the kernel Koopman eigendecomposition allows us to gain insight into the characteristic properties of the system by just analyzing a video of it.

\subsection{Cloud and aerosol simulation}

\begin{figure}[tbhp]
    \centering
    \begin{minipage}[t]{0.28\textwidth}
        \centering
        \subfiguretitle{(a) Frame 0}
        \includegraphics[width=\textwidth]{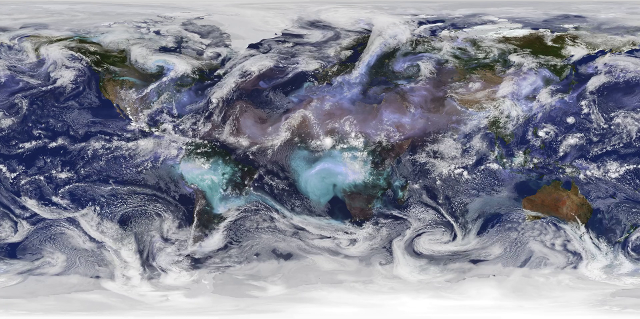} \\[-0.5ex]
        \subfiguretitle{(b) Frame 900}
        \includegraphics[width=\textwidth]{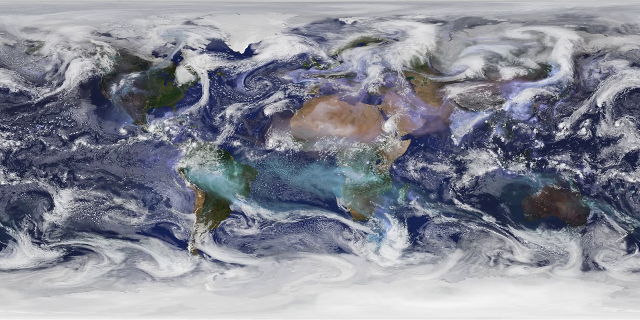}
    \end{minipage}
    \hspace*{0.2ex}
    \begin{minipage}[t]{0.4\textwidth}
        \centering
        \subfiguretitle{(e)}
        \includegraphics[width=\textwidth]{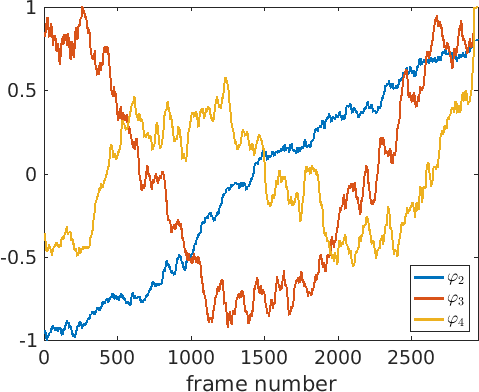}
    \end{minipage}
    \hspace*{0.2ex}
    \begin{minipage}[t]{0.28\textwidth}
        \centering
        \subfiguretitle{(c) Frame 1800}
        \includegraphics[width=\textwidth]{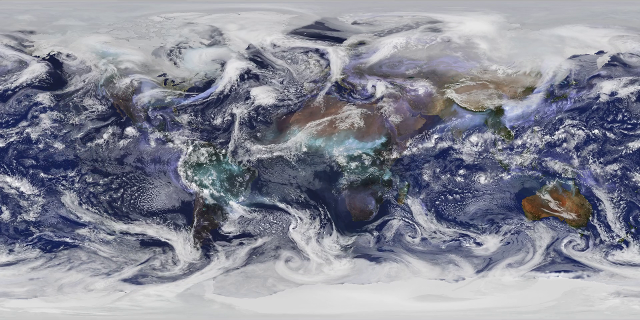} \\[-0.5ex]
        \subfiguretitle{(d) Frame 2700}
        \includegraphics[width=\textwidth]{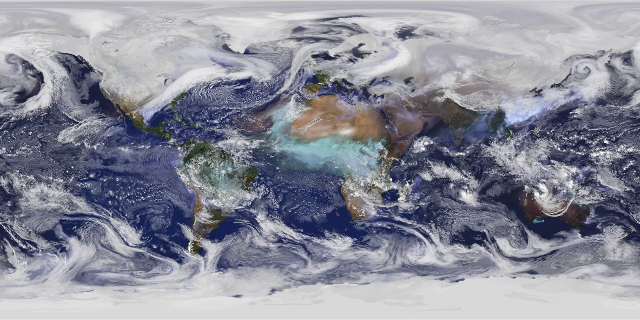}
    \end{minipage}
    \caption{(a--d) Frames of the video. (e) Dominant eigenfunctions evaluated at each frame.}
    \label{fig:Simulated clouds}
\end{figure}

\begin{figure}[tbhp]
    \centering
    \begin{minipage}{0.32\textwidth}
        \centering
        \subfiguretitle{(a) $ \lambda_2 \approx 0.98 $}
        \includegraphics[width=\textwidth]{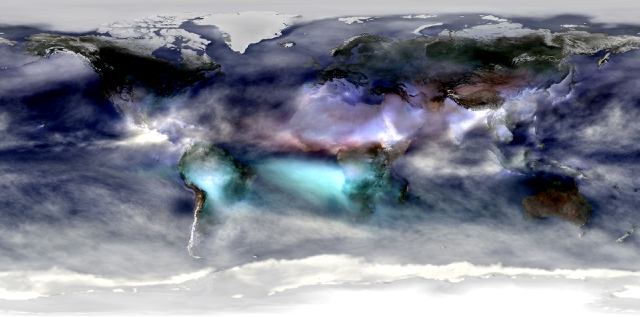} \\
        \includegraphics[width=\textwidth]{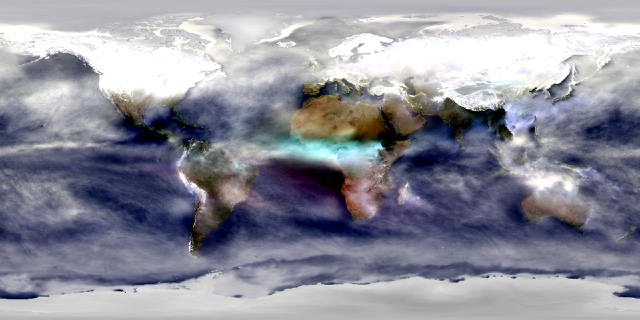}
     \end{minipage}
    \begin{minipage}{0.32\textwidth}
        \centering
        \subfiguretitle{(b) $ \lambda_3 \approx 0.96 $}
        \includegraphics[width=\textwidth]{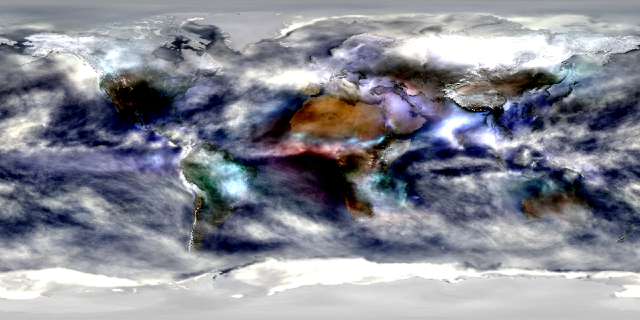} \\
        \includegraphics[width=\textwidth]{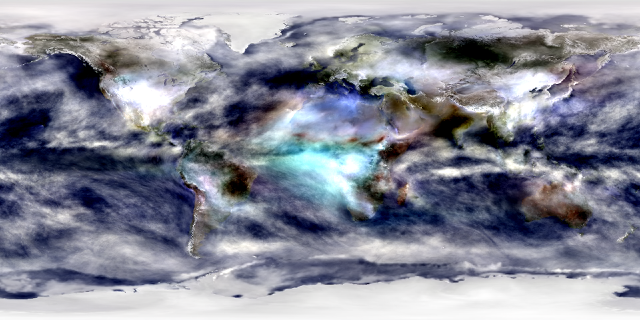}
    \end{minipage}
    \begin{minipage}{0.32\textwidth}
        \centering
        \subfiguretitle{(c) $ \lambda_4 \approx 0.93 $}
        \includegraphics[width=\textwidth]{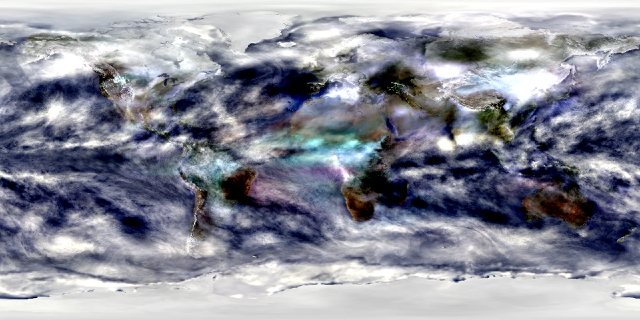} \\
        \includegraphics[width=\textwidth]{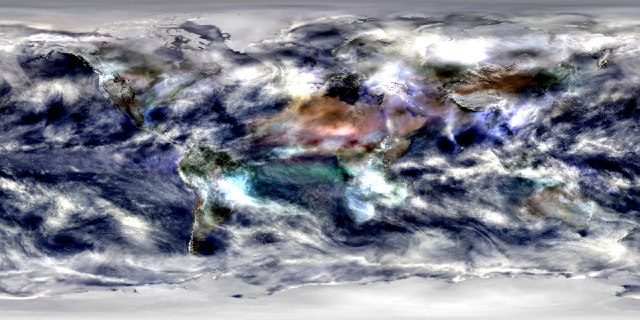}
    \end{minipage}
    \caption{Minima (top) and maxima (bottom) of the leading eigenfunctions of the kernel Koopman operator computed using gradient descent/ascent methods.}
    \label{fig:Simulated clouds eigenfunction maximizers}
\end{figure}

Let us now study a more complex dynamical system and apply the proposed approach to a video showing the simulation of clouds and aerosols generated by researchers from NASA's Global Modeling and Assimilation Office and NASA's Scientific Visualization Studio.\footnote{NASA: Simulated Clouds and Aerosols (\url{https://svs.gsfc.nasa.gov/30591}).} The video demonstrates how winds transport aerosols around the world; clouds are shown in white, dust in brown shades, sulfates in purple shades, and organic black carbon in green shades. The simulation starts on September 1, 2005 and ends on December 31, 2005 with hourly updates. The resolution of the video is 1080p, resulting in $ \mathbf{X}, \mathbf{Y} \in \R^{1080 \times 1920 \times 3 \times 2955} $. That is, the state space is 6220800-dimensional. We choose the same kernel as above, adjusting only the bandwidth $ \sigma $ and the regularization parameter $ \varepsilon $. Here, we choose $ \sigma = 1.25 \cdot 10^5 $ and $ \varepsilon = 1 $. The kernel Koopman eigendecomposition again results in several real-valued eigenvalues close to one corresponding to the slowest processes of the system. A few frames of the video and the (nontrivial) dominant eigenfunctions evaluated at each frame are shown in Figure~\ref{fig:Simulated clouds}.
We again apply the gradient descent and ascent techniques described above. The results are shown in Figure~\ref{fig:Simulated clouds eigenfunction maximizers}. The first nontrivial eigenfunction describes the slow passage of the aerosols over South America and the South Atlantic to Central Africa and the formation of clouds over North Australia. Furthermore, it contains information about the snow accumulation in parts of North America, Europe, and Russia. The eigenfunctions associated with the smaller eigenvalues pick up the faster and more local cloud and aerosol formations. Instead of applying the kernel-based methods to the video data, these methods could also be applied directly to the simulation data.

\subsection{Fluid dynamics problem}

\begin{figure}[tbhp]
    \centering
    \begin{minipage}{0.49\textwidth}
        \centering
        \subfiguretitle{(a)}
        \includegraphics[width=\textwidth]{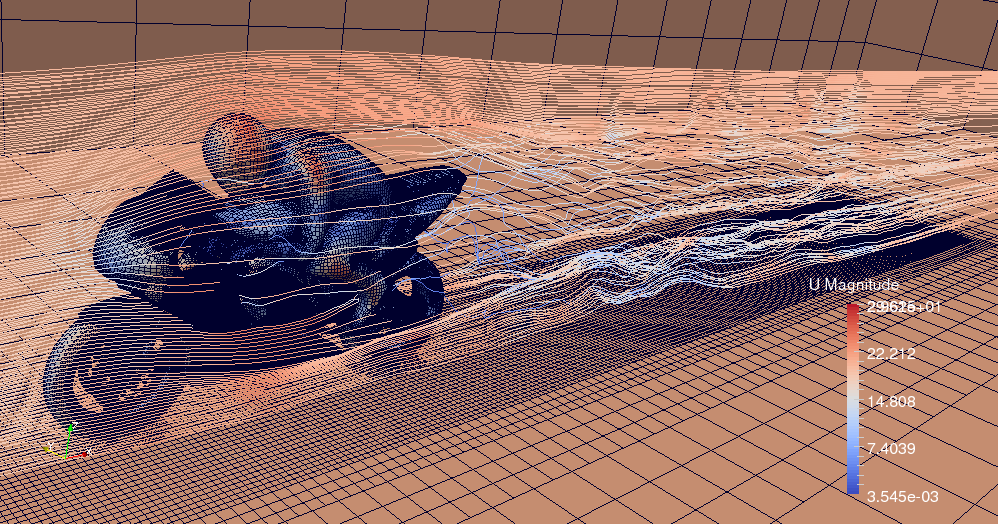}
    \end{minipage}
    \begin{minipage}{0.49\textwidth}
        \centering
        \subfiguretitle{(b)}
        \includegraphics[width=\textwidth]{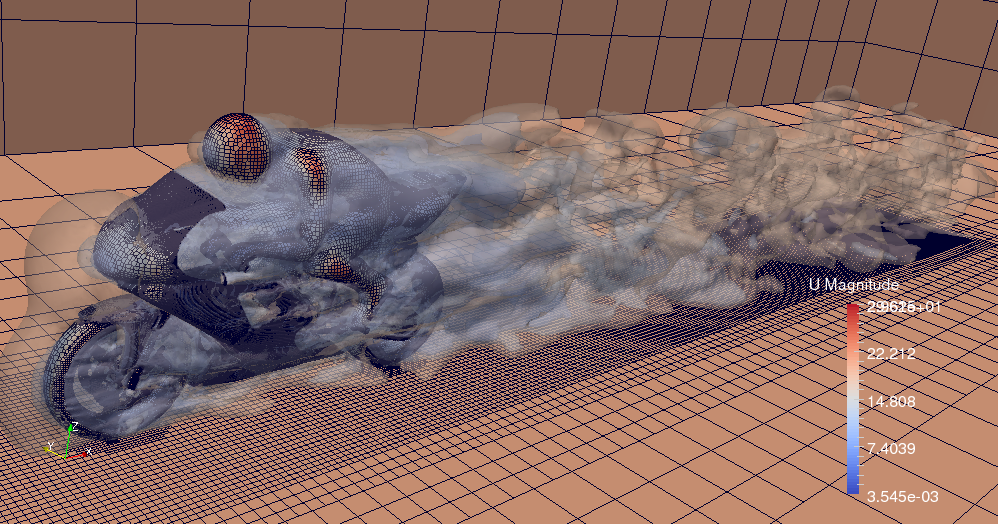}
    \end{minipage} \\[1ex]
    \begin{minipage}{0.49\textwidth}
        \centering
        \subfiguretitle{(c) $ \lambda_2 \approx 0.99 + 0.02 \ts \mathrm{i} $}
        \includegraphics[width=\textwidth]{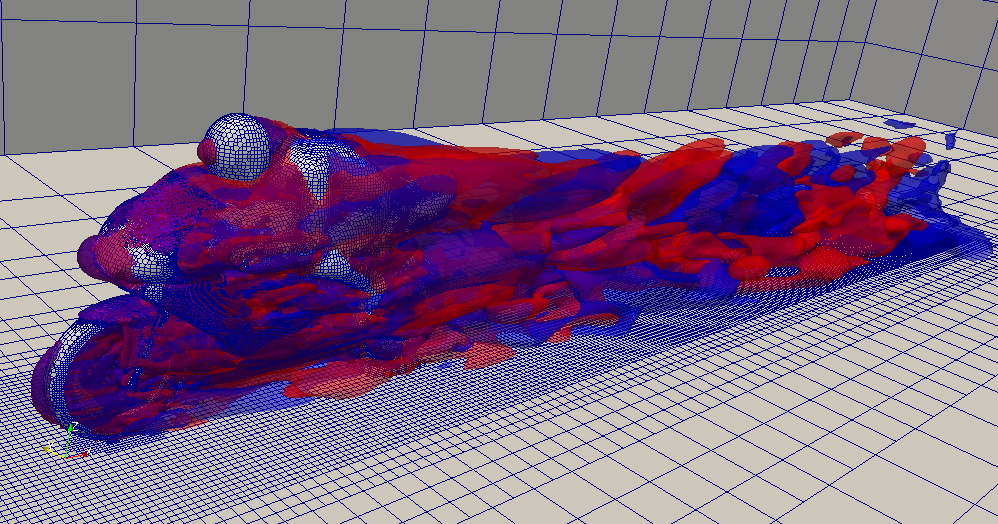}
    \end{minipage}
    \begin{minipage}{0.49\textwidth}
        \centering
        \subfiguretitle{(d) $ \lambda_4 \approx 0.98 + 0.03 \ts \mathrm{i} $}
        \includegraphics[width=\textwidth]{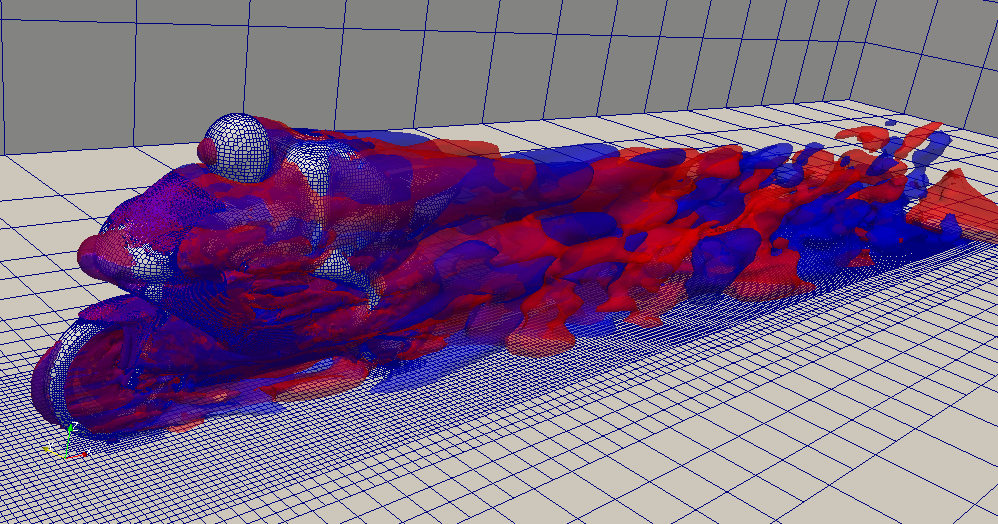}
    \end{minipage} \\[1ex]
    \begin{minipage}{0.49\textwidth}
        \centering
        \subfiguretitle{(e) $ \lambda_2 \approx 0.92 + 0.07 \ts \mathrm{i} $ }
        \includegraphics[width=\textwidth]{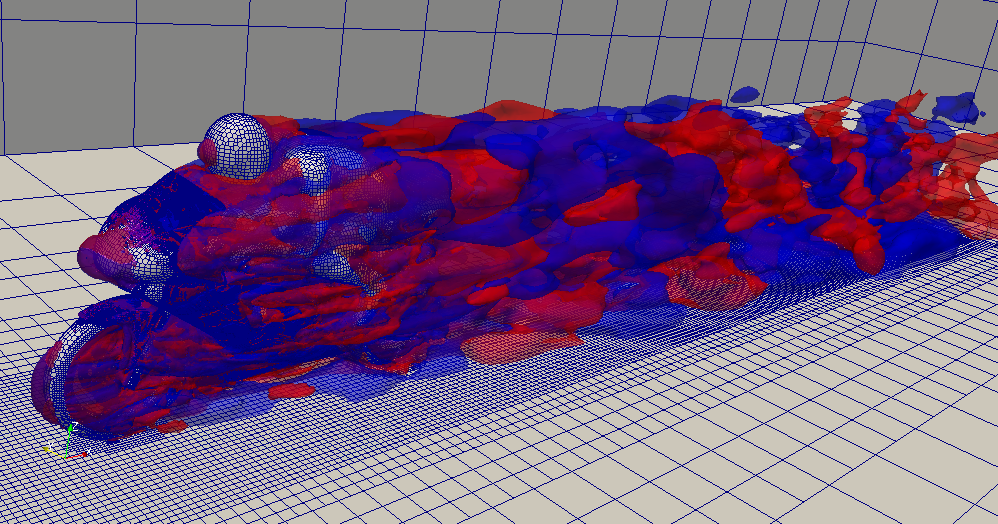}
    \end{minipage}
    \begin{minipage}{0.49\textwidth}
        \centering
        \subfiguretitle{(f) $ \lambda_6 \approx 0.91 + 0.14 \ts \mathrm{i} $}
        \includegraphics[width=\textwidth]{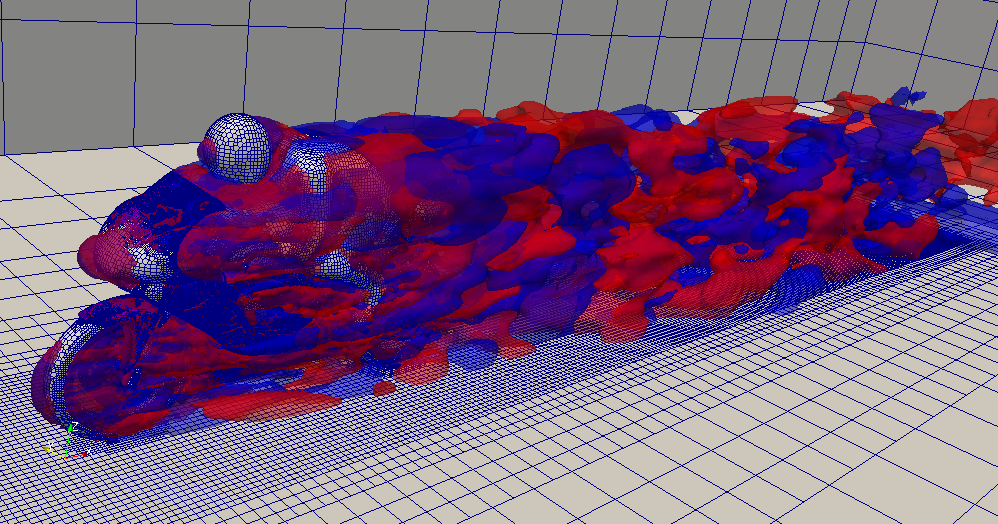}
    \end{minipage}
    \caption{(a) Model and surface mesh of the motorcycle problem. Streamlines and surface color according to the instantaneous velocity magnitude for $Re = 2.7 \cdot 10^6$. (b) Isosurface of the velocity field in (a). (c) Minimum (blue) and maximum (red) of the second eigenfunction at $Re = 2.7 \cdot 10^5$. (d) Fourth eigenfunction at $Re = 2.7 \cdot 10^5$. (e) Second eigenfunction at $Re = 2.7 \cdot 10^6$. (f) Sixth eigenfunction at $Re = 2.7 \cdot 10^6$.}
    \label{fig:Motorcycle}
\end{figure}

As a last example, we consider a fluid dynamics problem, the simulation of a driving motorcycle, which is part of the OpenFOAM package~\cite{JJT07}. Based on the motorcycle velocity $ U_\infty = 2\, \textrm{m}/\textrm{s}$ and $U_\infty = 20\, \textrm{m}/\textrm{s}$, respectively, its length $ L \approx 2\, \textrm{m} $, and the kinematic viscosity $\nu = 1.5\cdot 10^{-5}\, \textrm{m}^2/\textrm{s}$, the Reynolds number is $ Re \approx 2.7 \cdot 10^5 $ and $ Re \approx 2.7 \cdot 10^6 $, i.e., we are in a highly turbulent flow regime. A direct numerical simulation would significantly exceed today's computational capacities, which is why a turbulence model has to be used in order to reduce the number of grid cells. Here, we use a hybrid RANS/LES model, the delayed detached-eddy simulation (DDES), see \cite{SDS+06} for details. The numerical mesh is generated automatically using the OpenFOAM  utility \emph{snappyHexMesh}, which creates hexa-dominant meshes from triangulated surface geometries (e.g., CAD geometries), see Figure~\ref{fig:Motorcycle}(a) for the surface mesh around the motorcycle. The resulting mesh has approximately $3.2\cdot 10^6$ cells with highly non-uniform volumes. The \emph{PISO} scheme \cite{FP02} with a step size of $10^{-4}$ is used for the numerical solution.
We again utilize a Gaussian kernel for the computation of the eigenfunctions. Figures~\ref{fig:Motorcycle}(c) to (f) show the minima (blue) and maxima (red) of different kernel Koopman eigenfunctions at $Re = 2.7 \cdot 10^5$ and $Re = 2.7 \cdot 10^6$, respectively, where the left pictures correspond to the eigenvalues with a lower phase. For each computation, 2001 snapshots of the velocity magnitude with a lag time of $0.005\, \textrm{s}$ are used, i.e., $ \mathbf{X}, \mathbf{Y} \in \R^{3213298 \times 2000} $. In both cases, we observe that higher order modes such as (d) and (f) appear to possess structures of smaller scale which is consistent with standard DMD approaches, see, e.g., \cite{RMBSH09} for comparison. We also observe that this distinction becomes less evident with increasing Reynolds number. This emphasizes that scale separation becomes more difficult which is also in accordance with the physics of turbulent flows.

\section{Conclusion}

We have shown that the eigenfunctions of kernel transfer operators associated with high-dimensional systems contain valuable global information about the underlying dynamics and can be computed efficiently from time-series data. By choosing an appropriate kernel, the eigenfunctions can be estimated in a purely data-driven fashion, detailed knowledge about the system itself is not required. Selecting the kernel and tuning its hyperparameters could be automated using cross-validation techniques. This will be future work. Challenging new applications could be background subtraction or object tracking. Also predicting the next frame of a video or filling in missing frames could be future work. Furthermore, it would be interesting to systematically compare the coherent patterns obtained using optimized kernel Koopman eigenfunctions with standard DMD modes, which are frequently used for model reduction and control. DMD assumes linear dynamics, i.e., $ y_i = A \ts x_i $, whereas our approach implicitly takes into account infinitely many nonlinear basis functions for approximating the dynamics. In order to be able to apply DMD to high-dimensional fluid dynamics problems, 
due to the enormous size of the data sets, low-rank tensor approximation methods might be advantageous, see \cite{KGPS18}. Another open question is what kind of information the corresponding embedded transfer operators (e.g., the conditional mean embedding) and their eigenfunctions contain and how it can be exploited. Moreover, instead of using the eigenfunctions, analyzing left and right singular functions might be beneficial, in particular for understanding non-reversible dynamical systems.

\subsubsection*{Acknowledgments}

This research has been partially funded by Deutsche Forschungsgemeinschaft (DFG) through grant CRC 1114 \emph{``Scaling Cascades in Complex Systems''}. The OpenFOAM calculations were performed on resources provided by the Paderborn Center for Parallel Computing ($PC^2$).

\bibliographystyle{unsrt}
\bibliography{OKTOE}

\end{document}